\documentclass[manuscript,screen]{acmart}
\AtBeginDocument{%
  \providecommand\BibTeX{{%
    \normalfont B\kern-0.5em{\scshape i\kern-0.25em b}\kern-0.8em\TeX}}}

\setcopyright{acmcopyright}
\copyrightyear{2021}
\acmYear{2021}
\acmConference[KDD-MLF-2021]{KDD-MLF ’21}{August 14--18, 2021}{Virtual Workshop}
\acmBooktitle{KDD-MLF ’21, August 14--18, 2021, Virtual Workshop}
\acmPrice{15.00}
\acmISBN{978-1-4503-XXXX-X/YY/MM}

\usepackage{lipsum}
\usepackage{mwe}

\begin{document}

\title{Relational Graph Neural Networks for Fraud Detection in a Super-App environment}


\author{Jaime D. Acevedo-Viloria}
\email{jaime.acevedo@rappi.com}
\authornotemark[1]
\affiliation{%
  \institution{Rappi}
  \city{Bogotá}
  \country{Colombia}
}

\author{Luisa Roa}
\email{luisa.roa@rappi.com}
\affiliation{%
  \institution{Rappi}
  \city{Bogotá}
  \country{Colombia}
}

\author{Soji Adeshina}
\email{adeshina@berkeley.edu}
\affiliation{%
  \institution{University of California Berkeley}
  \city{Berkeley}
  \state{California}
  \country{USA}
}

\author{Cesar Charalla Olazo}
\email{cesar.charalla@rappi.com}
\affiliation{%
  \institution{Rappi}
  \city{Lima}
  \country{Per\'u}
}

\author{Andr\'es Rodr\'iguez-Rey}
\email{a3rodrig@ucsd.edu}
\affiliation{%
  \institution{University of California, San Diego}
  \city{La Jolla}
  \state{California}
  \country{USA}
}

\author{Jose Alberto Ramos}
\email{jose.ramos@rappi.com}
\affiliation{%
  \institution{Rappi}
  \city{Ciudad de M\'exico}
  \country{M\'exico}
}

\author{Alejandro Correa-Bahnsen}
\email{alejandro.correa@rappi.com}
\affiliation{%
  \institution{Rappi}
  \city{Bogotá}
  \country{Colombia}
}

\renewcommand{\shortauthors}{Acevedo-Viloria, et al.}

\begin{abstract}
Large digital platforms create environments where different types of user interactions are captured, these relationships offer a novel source of information for fraud detection problems. In this paper we propose a framework of relational graph convolutional networks methods for fraudulent behaviour prevention in the financial services of a Super-App. To this end, we apply the framework on different heterogeneous graphs of users, devices, and credit cards; and finally use an interpretability algorithm for graph neural networks to determine the most important relations to the classification task of the users. Our results show that there is an added value when considering models that take advantage of the alternative data of the Super-App and the interactions found in their high connectivity, further proofing how they can leverage that into better decisions and fraud detection strategies.
\end{abstract}



\begin{comment}
\begin{teaserfigure}
  \includegraphics[width=\textwidth]{sampleteaser}
  \caption{Seattle Mariners at Spring Training, 2010.}
  \Description{Enjoying the baseball game from the third-base
  seats. Ichiro Suzuki preparing to bat.}
  \label{fig:teaser}
\end{teaserfigure}
\end{comment}

\maketitle

\section{Introduction}

One of the main challenges for digital platforms is fraud detection/prevention, as it implies a cost; this cost materializes either economically or in terms of perception from negative outlooks on the user experience \cite{2020acfe}; according to 2020 ACFE report \cite{2020acfe} organizations lose 5\% of revenues each year. Particularly in digital platforms, the best known fraud modalities are account take over (ATO), credit card fraud and malicious accounts creation (bots). In these modalities, fraudsters take possession of legitimate accounts, steal cards, and create attacks of accounts created with bots to carry out unauthorized transactions and fraudulent activities.

An important stage of fraud detection is the early identification of fraudulent accounts, that is, users who enter the platforms with the aim of committing any type of malicious activity. Unlike malicious account bots, the creation of these accounts is not done in a massive way and fraudsters identify and exploit the vulnerability of the systems by creating accounts that meet the properties of legitimate accounts. By recognizing these types of users, the risk and losses are reduced while the fraud detection system is improved,  however these type of fraud entails great challenges as, at the best of our knowledge, few authors have studied it. 

Usually, platforms align machine learning models with business strategies in which a fraud expert establishes rules for the validation of different activities, an example of such a business strategy would be the requirement of biometrical account validation or email confirmation. Oftentimes this business strategies are supported by predictive models based on the historical data of the digital platform.  However, a common issue with this type of models is the lack of data for new accounts, this issue can be addressed by leveraging large platforms data like Super-Apps.

Super-Apps are mobile applications that integrate different functionalities in such a way that users can find a large variety of services in one place. This model originates with Chinese Super-Apps as WeChat and Alipay, that offer services ranging from delivery, leisure, and travel to financial services \cite{liu2017}, this model is also taking place in other regions with companies such as Grab, Mercado Libre, and Rappi. Given the nature of Super-Apps, different type of information, regularly referred to as alternative data, can be collected due to the variety of functionalities. Therefore, previous behavioral and consumption information of the users, as well as relational data of the users with entities as devices, credit cards, merchants, addresses and even users, may be of interest. Particularly, for functionalities related to financial services, such as a virtual wallet, this information gathering is of great importance as it allow to understand patterns and identify interactions that indicate possible fraudulent behavior of new financial services users. In the industry and literature, apart from variables based on transactionality and models like logic regression and decision trees \cite{abdallah2016fraud} are traditionally used, a new trend for fraud detection is the use of graph-based techniques.

Graphs have become one of the most studied areas due to the advantages and benefits they present. In particular, graphs allow capturing local and global information of an individual or object within a network, for fraud problems this represents a great opportunity since it is possible to understand patterns of malicious activities at a level that considers not only the specific data but also the relationships and context. For example, Roa et al. \cite{rao2020xfraud}, developed an explainable fraud transaction prediction system based on heterogeneous networks that allows defining a risky transaction score, likewise Wu and Chen \cite{wu2021connecting} propose a time risk control systems to detect collective fraud using graphs data bases.

In this paper we propose a framework for the comparison of different graph-based algorithms for possible fraud behaviour detection on newly registered users, where we determine the effectiveness of developing different multi-relational graphs to be modeled by a relational graph convolutional network approach. Furthermore, the utility of transactional variables before and after the registering process for the financial services of the Super-App will be evaluated in the defined methods. Finally, we will enhance the interpretability of the designed models using the GNNExplainer \cite{ying2019gnnexplainer} algorithm to identify the most important relationships. Four different experimental setups involving different timelines were drawn to prove the robustness of the measures obtained for each algorithm. As such, we aim to explore the value of the selected graph-based solutions to answer the following research questions:  
\begin{enumerate}
    \item How effective are GNN-based models for fraud prevention?
    \item What is the best GNN architecture for fraud prevention in a Super-App environment?
    \item What are the most important relationships in a Super-App based graph? %How graph-based approaches enhance a fraud detection system
    \item Does Super-App data improve the detection of fraudulent behaviour?
\end{enumerate}
The rest of the paper is organized as follows: Section~2 presents the related works along with the technical details of the machine learning models used in the paper; Section~3 introduce the alternative data that a Super-App entails; Section~4 describes the proposed methodology; Section~5 gives a detailed description of the experiments settings and characteristics; Section~6 presents the results; finally in Section~7 we present conclusions and future work.

\section{Background and related work}

When only transactional or behavioral data of an user is considered the environment and context of the information is lost; however graphs preserve this information while the behavior of the users is analyzed. One way to preserve this environment, or neighborhood, information is through centrality measures of the nodes, variables known as graph-based features. Said features have proven to be highly predictive in several fraud detection problems; Fakhraei et al. \cite{fakhraei2015collective}, found that by incorporating graph-based features like PageRank, k-core, Degree, Graph Coloring and Connected Components in a multi-relational graph the can improve spammer account classification performance. Likewise, Van Vlasselaer et al. \cite{van2017gotcha} studied the impact of network information in the context of social security fraud by using graph-based features from a time-weighted network and combing them with tabular data. Graph-based features have also been used in the context of camouflage behavior \cite{dou2020enhancing} proving to be of great value to assess fraud. 

Graphs also leveraged models known as Graph Neural Network (GNNs), are shown to be very effective at accumulating and encoding features from local neighborhoods \cite{Wu_2021}. Most of these models are designed for homogeneous graphs where only a type of node is considered, particularly for multi-relational structures the Relational Graph Convolution Network (RGCN) \cite{schlichtkrull2018modeling} provides a framework where the structure of the Graph Convolutional Networks (GCN) is extended to heterogeneous graphs. The applications of these models are extensive, specifically for fraud, Liu et al. \cite{liu2018heterogeneous}, for instance, studied malicious account detection in the world's leading mobile payment platform with graph neural networks that modeled the way devices and accounts aggregated in the ecosystem of Alipay.

%\subsection{XGBoost}
%Extreme Gradient Boosting, better known as XGBoost, is a boosting tree based supervised machine learning algorithm \citet{chen2016xgboost}. Extremely popular for its high efficiency, scalability, and prediction effectiveness; XGBoost has become on of the standards classification algorithms used for tabular data. The algorithm, is an ensemble of sequentially built boosting decision trees, where each sub-model learns from previous errors to enhance the prediction. The boosting its done via Gradient descent, calculating error or residuals from an objective function that measures the errors between predictions and labels.

\subsection{Graph Neural Networks}
In the last couple of years one of the key methods for classification problems dealing with non-euclidean data, particularly data coming from networks, has been that of Graph Neural Networks \cite{Wu_2021}. For this paper we will implement a method proposed by Schlichtkrull et al. \cite{schlichtkrull2018modeling} for classification problems for multigraphs arising from knowledge bases. For this, we strat by defining a heterogeneous graph $G= (V, E, R, Y)$  to be a set of nodes $V$, directed edges $E$ and relationship types $R$, together with a set of node labels $Y$. Here every edge in $E$ is described as a triple $(v_i, r, v_j)$ where $v_i, v_j\in V$ and $r\in R$.  
\subsubsection{Relational Graph  Convolutional Networks}
A Relational Graph Convolutional Neural Network (R-GCN) \cite{schlichtkrull2018modeling} consists of a simple propagation model that computes a forward-pass update of a node $v_i\in V$ in a heterogeneous network by the rule
\[
h_i^{(l+1)}= \sigma \left( W_0^{(l)}h_i^{(l)}+\sum_{r\in R}\sum_{j\in \mathcal{N}_i^r} \frac{1}{c_{i,r}}W_r^{(l)}h_j^{(l)}\right),
\]
where $\mathcal{N}_i^r$ denotes the set of neighbor indices of $v_i$ under the relation $r$, $c_{i,r}= |\mathcal{N}_i^r|$, $W_r^{(l)}$ is a trainable weight matrix in the $l$-th layer, and $\sigma$ is an activation function. Here $W_r^{(l)}$ is modeled to be a convolutional layer (GCN) of a homogeneous graph neural network as in Kipf et al. in \cite{GCN}, while being regularized by what Schlichtkrull et al. call the basis and block diagonal decompositions. Essentially, at each layer, the network is solving the classification problem by running a GCN on each of the relational components and aggregating them together. It does this by means of minimizing the loss given by 
\[
\mathcal{L}=-\sum_{i\in Y}\sum_{k=1}^K t_{ik} \log h_{ik}^{(L)}
\]
where $h_{ik}^{(L)}$ is the $k$-th entry of the network output for the $i$-th laveled node, and $t_{ik}$ denotes its respective ground truth label. 

\subsubsection{Explainability}
Given the high complexity of GNNs, the interpretation of the predictions is one of the most challeging tasks. Recently, different techniques have been developed for intepretability, among them, GNNExplainer \cite{ying2019gnnexplainer} an agnostic approach that provides interpretable explanations for any GNN task by using nodes and features masks to select important inputs. For this, first a subgraph $G_s$ is defined that starts from the complete graph $G_c$, with this the nodes and features masks are randomly initialized. Then, the masks are combined with the complete graph through element-wise multiplications and optimized by maximizing the mutual information ($MI$) between the predictions of the complete graph and subgraph with
\[
\max_{G_s} MI (Y,(G_s,X_s)) = H(Y) - H(Y|G=G_s,X=X_s)
\]
Where $MI$ quantifies the change in the probability of prediction considering a subgraph $G_s$ and the associated features $X_s$. This change is calculated by making the difference between the fixed entropy of the prediction of the trained GNN ($H(Y)$) and entropy of the prediction given $G_s$ and $X_s$, also, due to $H(Y)$ being a constant this optimization problem can be taken as the minimization of the conditional entropy given $G_s$ and $X_s$. Consequently, the node mask selects the important relationships of the subgraph and the feature mask identifies the variables with the most information.

\section{Super-App Alternative Data}

As fraud becomes more complex and fraudsters find gaps in traditional prevention systems, hence anti-fraud techniques must combine alternative data and machine learning to provide robust solutions \cite{Hamilton2020}. Specifically for new users of financial services, alternative data is defined as information that cannot be captured through traditional sources such as credit bureaus or the application process.

According to Michele Tucci, CredoLab Chief Product Officer \cite{GBG2020}, alternative data for fraud detection can be divided into two large groups: transactional and behavioral. The first group includes data payment records of utilities, telecom data, e-commerce data, among others. This source of information allows us to capture the transactionality of individuals in different settings in such a way that patterns can be defined. The second group consists of behavioral information such as travel patterns, e-commerce activity, social media data \cite{zou2020differential} and may even consider personal information such as criminal records, and employment history. In such a way that personal information can be captured as well as activities that include interactions and relationships with other entities.

Particularly, given the great variety of functionalities and the ecosystem that a Super-App entails, different transactional an behavioral alternative data can be obtained from these applications. \cite{roa2021super} defines variables that add different measures over time on consumption in delivery, transportation and financial services functionalities such as amount spent, number of orders, average tip, use of discounts, number of money transfers, among others. The authors also define generic variables in which demographic information of the user is captured, such as age, place of residence, number of addresses, number of credit cards registered and others. Furthermore, the interactions within the Super-App also become a non-traditional source of information which can be translated into graphs.

\begin{figure}
    \centering
    \begin{minipage}{0.45\textwidth}
        \centering
        \includegraphics[scale = 0.30]{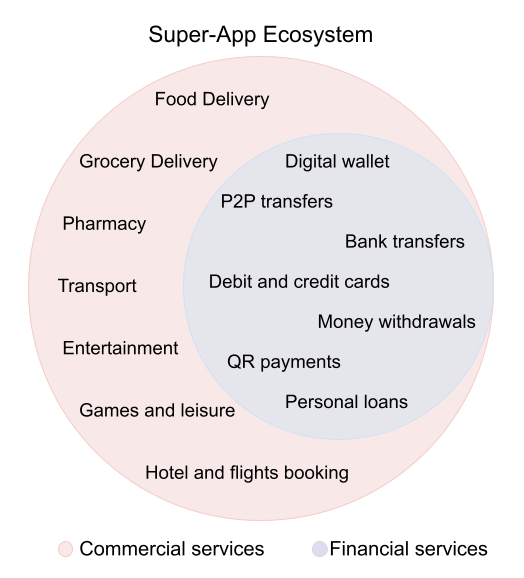}
        \caption{Functionalities and users}
        \label{fig:diagram}
    \end{minipage}\hfill
    \begin{minipage}{0.45\textwidth}
        \centering
        \includegraphics[scale=0.45]{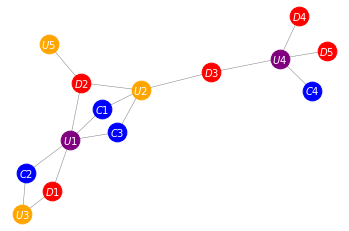}
        \caption{Graph with financial users (violet), only non-financial users (orange), devices (red) and credit cards (blue).}
        \label{fig:graph}
    \end{minipage}
\end{figure}

The functionalities of a Super-App can be separated into \textit{commercial} and \textit{financial} services.  As shown in Figure \ref{fig:diagram}, commercial services includes all functionalities related to restaurants, supermarkets, traveling, among others while financial services include all related to payments, transfers and different loans. Similarly, the customers are also differentiated as \textit{non-financial services} and \textit{financial services} users, where both have interacted with commercial services but only the latter have acquired a financial product. Additionally, for the financial services in particular, Super-Apps must make different validations before granting the product to the user, for example, for loans and credits, a prior credit evaluation must be done to assess the risk.

%For this paper we are differentiating the services of the Super App as the commercial services, which includes restaurants, supermarkets, traveling, among others; and the financial services which requires further validation compared to the commercial services. Similarly, the customers are also differentiated as \textit{financial services} and \textit{non-financial services} users, where both groups have interacted with the services as shown in Figure \ref{fig:diagram}. BAJAR --> %Therefore, both set of users have historic commercial services transactional and behavioral features (\textit{historical features}), while financial users additionally have transactional features from the commercial services after the validation for the financial services (\textit{post-validation features}). 

%\begin{figure}
%    \centering
%    \includegraphics[scale = 0.35]{images/Copia de ven_.png}
%    \caption{Functionalities and users}
%    \label{fig:diagram}
%\end{figure}

%----------------------------------------------------------------------------------------------------------------------------------
\section{RGCNs for Fraud Detection in Super-App environment} % Relational Graph Neural Networks for Fraud Detection

In this section we propose a method that allows to infer whether a newly registered financial services user is going to commit fraudulent behavior. First, since there is a validation process for financial products, we define what we call \textit{historical features} and the \textit{post-validation features}. These two types of features are available to financial users whereas for non-financial users only historical features are available. Furthermore, to exploit the advantages given by the information of relationships among our users we constructed a graph whose vertices are the users, devices and credit cards; and whose edges relate users with their devices used for login, as well as users with their registered credit cards. We illustrate an example graph in Figure \ref{fig:graph}. Specifically, these relationships become of great value as these capture abnormal behaviors in the use and registration of devices and credit cards even before a user makes a transaction.

%\begin{figure}
%    \centering
%    \includegraphics[scale=0.5]{images/GRAPH_EXAMPLE.png}
%    \caption{Graph with financial users (violet), only non-financial %users (orange), devices (red) and credit cards (blue).}
%    \label{fig:graph}
%\end{figure}

Therefore we create a framework for the comparison of different models, using the information captured by these graphs with transactional and behavioral data of our users, in a set of heterogeneous graph neural network models for fraud prediction. Particularly, we proposed the following experiments:
\begin{enumerate}
    %\item \textit{Transactional and Behavioral Data + Graph Based Features}: Classification model for financial users that considers historical features, post-validation features and graph-based features obtained from the previously defined network. %The graph-based features are the node's degree, Pagerank centrality, eigenvector centrality and the number of neighbor user nodes that have already been classified as fraud. %we used an XGBoost classification model that considers the respective train and test divisions of the aforementioned datasets. In this experiment, Super-App transactional and Behavioral features before and after the wallet registration are considered for the new wallet users, in addition to the aforementioned graph-based features. %In addition, in this experiment all the features mentioned above are considered with a total of 291 per observation.
    \item \textit{RGCN 1}: RGCN model with no Super-App features in any of the nodes. The features to be used in this experiment are built by doing a random initialized embedding for every node of the network, where this embedding is further refined by the training of the neural network. % For these experiments the no-wallet users are consider as part of the train set. size 16
    \item \textit{RGCN 2}: RGCN model where historical and post-validation features are considered only for financial user nodes, for the other node types we use the same randomly initialized embedding of RGCN 1 experiment that is refined in the optimization.
    %before and after wallet registration marketplace features to the new user nodes, for the others node types we use the same randomly initialized embedding of RGCN 1 experiment.
    % that is an input size of 285, 
    \item \textit{RGCN 3}: RGCN model where for non-financial user nodes only historical features are considered, the financial users keep the historical and post-validation features. Finally, the other nodes receive the same randomly initialized embedding of experiment RGCN1 that is refined in the optimization.
    %we added the the marketplace before wallet registration to the no-wallet users nodes, the new users are kept with the complete market place features, the other nodes receive the same randomly initialized embedding of experiment RGCN1.     % input size of 138, and keep the new users with the complete market place features and the other nodes with the embedding of experiment RGCN1.
    \item \textit{RGCN 4}: RGCN model where for non-financial user nodes historical features are concatenated with a random initialize embedding that is refined in the optimization so the complete size has the same dimension as the financial user node. The rest of the nodes have the same features presented in experiment RGCN 3.
\end{enumerate}

Finally, the performance metric used to compare the results of the models is the Area Under the ROC Curve (AUC) as it indicates the ability of the model to discriminate the classes in different thresholds, and it is not affected as much by the highly unbalanced nature of the data. Furthermore, we use a GNNExplainer to identify the relationships that are more important for fraud detection, this is done by searching a representative node for the financial users in the network.

\section{Experimental Setup}

\begin{figure}[tbp!]
  \centering
  \includegraphics[scale = 0.7]{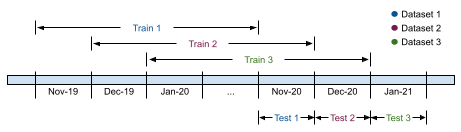}
  \caption{Datasets time frame}
  \label{fig:timeline}
\end{figure}

Our sample of interest consists of those new financial users who have confirmed their identity with an identification scanning and validation tool. Given that this validation requirement became mandatory in November 2019, we decided to  focus on those users who were approved for wallet use from November 2019 to February 2021. From this, three different datasets were defined, each one considering the registered and validated users over a time period of a year; Figure~\ref{fig:timeline}  and Table~\ref{tab:data_specifications} present a detailed description of these datasets. For each dataset we considered the network of users and their respective devices and credit cards. In these networks, all devices and credit cards that have been used by new financial users in the commercial services are considered, also non-financial users who have used the same devices or cards are included. Table \ref{tab:graphs_specifications} presents a summary of the details.

\begin{table}
\centering
  \caption{Datasets specifications}
  \label{tab:data_specifications}
  \begin{tabular} {p{\dimexpr.28\linewidth-1\tabcolsep}p{\dimexpr.15\linewidth-1\tabcolsep}p{\dimexpr.15\linewidth-1\tabcolsep}p{\dimexpr.203\linewidth-1\tabcolsep}p{\dimexpr.203\linewidth-1\tabcolsep}}
  %{rcccc}
    \hline  %cambio
    Time Frame            & Train Users & Test Users & Fraud Train (\%) & Fraud Test (\%)  \\
    \hline
    Nov. 2019 - Nov. 2020 & 55,411      & 3,496      & 6.46\%                 & 3.74\% \\
    Dec. 2019 - Dec. 2020 & 56,638      & 3,525      & 6.35\%                 & 3.69\% \\
    Jan. 2020 - Jan. 2021 & 59,983      & 8,033      & 6.14\%                 & 3.23\% \\
    \hline
  \end{tabular} % jajaj no puede ser, estabamos pensando en lo mismo como toca poner esto jaja
\end{table}

\begin{table}
\centering
  \caption{Graphs specifications}
  \label{tab:graphs_specifications}
  \begin{tabular} {p{\dimexpr.28\linewidth-1\tabcolsep}p{\dimexpr.15\linewidth-1\tabcolsep}p{\dimexpr.16\linewidth-1\tabcolsep}p{\dimexpr.15\linewidth-1\tabcolsep}p{\dimexpr.15\linewidth-1\tabcolsep}}
  %{rcccc}
    \hline
    Time Frame  & Users & Credit Cards & Devices & Total edges \\
    \hline
    Nov. 2019 - Nov. 2020 & 239.079    & 549.963          & 284.527      &1.175.428  \\
    Dec. 2019 - Dec. 2020 & 238.048    & 550.829          & 288.176      &1.183.316  \\
    Jan. 2020 - Jan. 2021 & 254.750    & 589.704          & 307.416      &1.259.951  \\
    \hline
  \end{tabular}
\end{table}

In our experiments, for the historical features we built more than 162 features such as amount spent, frequency of consumption, most used payment methods, preferences, use of discounts, among others; in post-validation features we included 138 variables that consider the average consumption, standard deviation and frequency of the first 5, 10 and 15 orders.

%Three types of features were considered in our experiments: transactional and behavioral data from the commercial services before and after the registering process for the financial services of the Super-App, and graph-based features from the network of interest. For the first group of features we are trying to collect enough information about the transactional data inside of commercial services in the Super-App, 

Considering the influx of new financial users of the Super-App we decided to train our models, mentioned in the previous section, with the first 11 months and test in the last month of the one year time period. Each of these experiments is evaluated on the three datasets. All the experiments use the same network structure, we consider 2 hidden layers of size 16 with Relu activation functions and a last layer with an output size of 2. We train these models using a cross entropy loss function weighted by the proportion of the minority class, given the imbalance of our data, with a learning rate of 0.01 and 10 epochs. Additionally, due to the size of the graphs we decided to do mini-batch training with a neighbour sampling, that is 15 edges are randomly selected for each node and a batch size of 4096. Finally, we compare the performance for every model on each dataset using the AUC. % {\color{orange}and it is not affected as much by the highly unbalanced nature of the data}.

For the GNN explainer, a representative node was found for the financial users in the network, for this case we take the embedding of all the users after the last projection layer and compute the embedding average, to then find the closest node to the average in terms of euclidean distance. Finally, for the selected node we compute the importance of each relation up to two hops away from the representative node, and then aggregate the importance of each node type around to determine which relations are most important for the output of the model.

\section{Results and Discussion}

The results for all the methods on each dataset are compared by their AUC metric, which is reported on Table \ref{tab:res}. 

%Overall, the results for the XGBoost consistently outperform every other model for each dataset, highlighting the effectiveness of this model, and pointing out toward the importance of user data in this type of experiment. To better understand which features of the user were most relevant we computed their SHAP values to determine the feature importance on each dataset. A dataset 3 SHAP values Figure \ref{fig:shap} is shown in representation of the three experiments due to the similarities of the results when computing the SHAP values for the features. All three experiments share the same first three features in SHAP value importance, highlighting the effectiveness of the canceled orders in the marketplace after the activation of the wallet, the historical canceled orders by the fraud detection system, and the eigenvector centrality of the created graph; other features are also shared as some of the most important on all three datasets, like the recency of the user ordering from a restaurant in the marketplace, the reactivation score determining how likely the user is to reactivate on the marketplace, and the maturity of the user in the marketplace.  

%\begin{figure*}
  %\centering
  %\includegraphics[scale = 0.40]{images/SHAP_exp3.png}
  %\caption{SHAP values for tabular experiment with dataset 3}
  %\label{fig:shap}
%\end
From the results we can see that the RGCN1 variant is the lowest performer of all the variations in every dataset, this behaviour of the model highlights the usefulness of the Super App features we are adding in the other models, and suggests that using graph structure with node features is the better approach. 

The RGCN3 has the best performance out of all the variants for all the datasets, this makes sense because this model has the most information, having features for both financial and non-financial users. It is also interesting to see the RGCN4 variant performance always be in the middle of RGCN2 and RGCN4, this makes sense because in essence, RGCN4 is a combination of both of those models. It can also be concluded that completing the feature set of the non-financial users with the learned embedding doesn't perform as well as having the feature set by itself with a different dimensionality. Finally, the RGCN2 variant is the worst performing of every model with features, this highlights that adding the feature set for the non-financial users is also augmenting the predictive power of the model.

%It is interesting that the XGBoost model works better than the RGCN variants, this can be attributed to the high effectivity of the variables that might be diluted while being aggregated over an RGCN methodology.

Additionally,  the results consistently decrease for each model as we use more recent datasets. This could be due to to the most recent transactions not being completely labeled as possible fraudster transactions, this can also be inferred when looking at the lower fraud percentages for the most recent dataset test set in Table \ref{tab:data_specifications}, where the fraud percentage is around 0.5\% less than the other two datasets.

%Update of results considering 5 seed me oresent the mean and the standard deviation \tabcolsep
\begin{table}
\centering
  \caption{AUC mean and standard deviation}
  \label{tab:res}
  \begin{tabular}{p{\dimexpr.15\linewidth-1\tabcolsep}p{\dimexpr.20\linewidth-1\tabcolsep}p{\dimexpr.20\linewidth-1\tabcolsep}p{\dimexpr.20\linewidth-1\tabcolsep}p{\dimexpr.20\linewidth-1\tabcolsep}}
  %\begin{tabularx}{\textwidth}
    \hline
    Dataset & RGCN1 & RGCN2 & RGCN3 & RGCN4 \\
    \hline
    Dataset 1 & 0,6538$\pm$0,0074 & 0.7988$\pm$0.0090 & 0.8021$\pm$0.0122 & 0.7962$\pm$0.0018\\
    Dataset 2 & 0,5950$\pm$0,0115 & 0.7673$\pm$0.0110 & 0.7747$\pm$0.0048 & 0.7629$\pm$0.0059\\
    Dataset 3 & 0,5626$\pm$0,0075 & 0.7452$\pm$0.0114 & 0.7536$\pm$0.0091 & 0.7460$\pm$0.0324\\
    \hline
  %\end{tabularx}
  \end{tabular}
\end{table}

Furthermore, the GNNExplainer algorithm was used to describe the importance of connections to the calculated fraud probability of the financial users. To this end a representative node of the network was found and used to compute the importance of each set of relations up to two hops away, these results are reported in table \ref{tab:explainer}.

With the GNNExplainer results we can see a clear progression to the changes made in the algorithms, for RGCN1 no node has features and therefore the importance of all other node types are very similar. Once we add features to the financial users in RGCN2 the devices seem to add the most information to those features. The, when adding features to the non-financial users for RGCN3, we can see those users taking more importance in the output while the device importance is diminished. However, for RGCN4 the concatenation of the randomly initialized embedding to the features diminishes the information provided by those relationships, and this can be seen in the Users close to zero importance.

%and RGCN4, we can see the non-financial users taking more importance in the output while the device importance is diminished.

\begin{table}
\centering
  \caption{Importance of each type of interaction for a representative node in Dataset 3 and each experiment.}
  \label{tab:explainer}
  \begin{tabular}{rccccc}
    \hline
    Relationship & RGCN1 & RGCN2 & RGCN3 & RGCN4 \\
    \hline
    Credit Cards &2.8282 &0.5195 &2.7977 &1.6792 \\
    Devices      &2.2261 &4.0970 &0.1048 &0.6880 \\
    Users        &2.2932 &1.2248 &8.3703 &$\epsilon \ll 0$ \\
    \hline
  \end{tabular}
\end{table}

\section{Conclusions}

Through the developed framework we determine a robust methodology for the comparison of distinct RGCN model, where the results obtained proof the effectiveness of these GNN methodologies for fraud detection.
As for the recommended architecture, the one used for RGCN3 is the more robust model, having the top AUC Score for all three datasets, and also because from the GNNExplainer we can conclude that the importance values of the Super-App information from the users relationships is being effectively taken.

We also find out that Super-App data is highly predictive and significantly improves the performance of our models, highlighted by the difference in performance of the RGCN1 in comparison to the other methodologies that do involve features obtained from the Super-App. We also discover from the difference in performances of RGCN3 and RGCN4 that concatenating this partial feature set of the non-financial users, with learnt embeddings from the optimization of the GNN to match the dimension of the financial users is not as effective as having the partial feature set by itself.

Additionally, we find out the most important relations for each of the implemented RGCN models, from which we can conclude that when not using features for the non-financial users the structural information given by the device connection is the most important, this is the case of RGCN2. But, when adding features to those non-financial users the structural information given by those users relations takes over the information given by the devices, becoming the most important relations for RGCN3. Finally, for the RGCN4 the concatenation is not as effective, proved by the results where the information taken from the Users relations becomes practically none.

For future work we would like to develop a model that takes into account that some of the features of different node types are shared, and therefore the should be learnt parallel between those different types on the coinciding features. Also, it could be interesting to evaluate the value of assigning features to non-user nodes in order to understand how fraud prediction is enhanced.

%\section{Acknowledgments}

%We want to thank George Karypis, Karthik Bharathy, Ian Robinson and the Neptune AWS team for their insightful ideas and valuable discussions.

\bibliographystyle{ACM-Reference-Format}
\bibliography{sample-base.bib}
%\nocite*{}

\end{document}